\documentclass[11pt]{article}

\usepackage[final]{acl}

\usepackage{times}
\usepackage{latexsym}

\usepackage[T1]{fontenc}

\usepackage[utf8]{inputenc}

\usepackage{microtype}

\usepackage{inconsolata}

\usepackage{graphicx}

\usepackage{booktabs}
\usepackage{amsmath}
\usepackage{tikz}
\usepackage{multirow}
\usepackage{pgfplots}
\usepackage{amssymb}
\usepackage{xcolor}
\pgfplotsset{compat=1.18}
\usetikzlibrary{arrows.meta, positioning, shapes.geometric, fit, backgrounds, calc}

\title{Diagnosing the Fact-Grounding Gap in Multi-Hop Question Answering}
\newcommand{\aspace}{\hspace{0.7em}}
\author{
  Kevin Mo$^{1}$ \aspace
  Nathan Mo$^{2}$ \aspace
  Richard Zhu$^{1}$ \\
  $^{1}$Independent \aspace
  $^{2}$Northwestern University
}

\begin{document}
\maketitle

\begin{abstract}
Multi-hop question answering requires combining information from multiple documents to answer complex questions. These systems have grown increasingly capable, yet when they fail, the error is typically attributed to not finding the right documents. Whether this holds at the level of individual reasoning steps remains largely unexamined. We investigate this across three standard multi-hop QA benchmarks and find that failures decompose into two distinct modes: \emph{retrieval failures}, where the needed passage was not retrieved, and \emph{extraction failures}, where the passage was retrieved but the needed fact could not be extracted --- a phenomenon we term the \emph{fact-grounding gap}. Extraction failures account for nearly half of all per-hop deficiencies and are invisible to standard retrieval metrics. They remain unresolved by every retrieval intervention we test, establishing a ceiling for retrieval-only improvements. The gap's severity varies across benchmarks and question types, but extraction failures appear on every dataset we measure. Our findings reveal that retrieval failures and extraction failures are fundamentally different bottlenecks requiring different solutions --- a distinction absent from current evaluation practice.\footnote{Code is available on GitHub: \url{https://github.com/RichardZhu123/fact-grounding-gap}}
\end{abstract}

\section{Introduction}

Multi-hop question answering requires combining information from multiple documents to answer complex questions, and remains a persistent challenge in NLP \citep{yang2018hotpotqa, trivedi2022musique, ho2020constructing}. A core difficulty is that each reasoning step depends on the previous one: if the system retrieves incorrect or insufficient evidence at any hop, subsequent steps propagate the error, leading to an incorrect final answer \citep{press2023measuring, trivedi2023interleaving}. This has motivated substantial work on improving how these systems retrieve evidence, including dense retrieval \citep{karpukhin2020dense}, adaptive strategies that decide when to search \citep{asai2024selfrag, jiang2023active}, and iterative retrieval-reasoning loops \citep{trivedi2023interleaving, yao2023react}. These approaches share a common assumption: retrieval quality is the primary bottleneck, and surfacing the right documents will lead to correct answers.

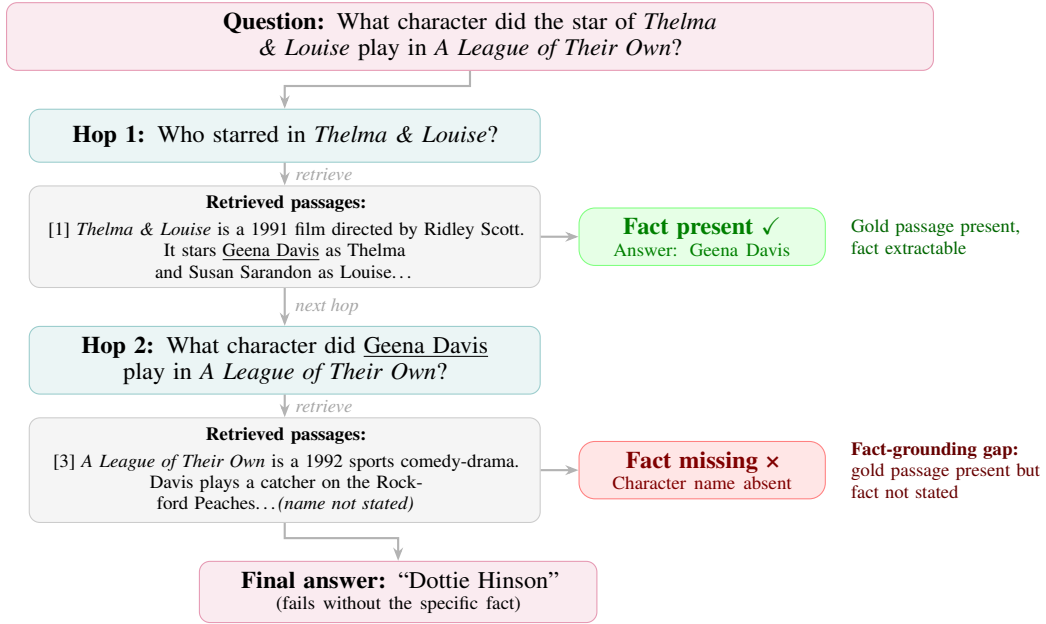
\begin{figure*}[t]
\centering
\begin{tikzpicture}[
    node distance=0.4cm and 0.3cm,
    box/.style={draw, rounded corners=4pt, minimum height=0.7cm, text width=#1, align=center, font=\small},
    box/.default=5cm,
    question/.style={box=12cm, fill=purple!8, draw=purple!40},
    subq/.style={box=6.5cm, fill=teal!8, draw=teal!40},
    passage/.style={box=6.5cm, fill=gray!8, draw=gray!40, minimum height=1.2cm, font=\scriptsize},
    verdict/.style={box=3cm, rounded corners=6pt, font=\small\bfseries},
    good/.style={verdict, fill=green!10, draw=green!50, text=green!50!black},
    bad/.style={verdict, fill=red!10, draw=red!50, text=red!50!black},
    answer/.style={box=5cm, fill=purple!8, draw=purple!40},
    arr/.style={-{Stealth[length=5pt]}, thick, gray!60},
    lbl/.style={font=\scriptsize\itshape, text=gray!70},
]

\node[question] (q) {\textbf{Question:} What character did the star of \textit{Thelma \& Louise} play in \textit{A League of Their Own}?};

\node[subq, below=0.5cm of q.south west, anchor=north west, xshift=0.3cm] (sq1) {\textbf{Hop 1:} Who starred in \textit{Thelma \& Louise}?};

\node[passage, below=0.3cm of sq1] (p1) {%
\textbf{Retrieved passages:}\\[2pt]
{[1]} \textit{Thelma \& Louise} is a 1991 film directed by Ridley Scott.\\
It stars \underline{Geena Davis} as Thelma and Susan Sarandon as Louise\ldots
};

\node[good, right=0.5cm of p1] (v1) {Fact present \checkmark\\[-2pt] {\scriptsize\normalfont Answer: Geena Davis}};

\node[subq, below=0.5cm of p1] (sq2) {\textbf{Hop 2:} What character did \underline{Geena Davis} play in \textit{A League of Their Own}?};

\node[passage, below=0.3cm of sq2] (p2) {%
\textbf{Retrieved passages:}\\[2pt]
{[3]} \textit{A League of Their Own} is a 1992 sports comedy-drama.\\
Davis plays a catcher on the Rockford Peaches\ldots \textit{(name not stated)}
};

\node[bad, right=0.5cm of p2] (v2) {Fact missing \texttimes\\[-2pt] {\scriptsize\normalfont Character name absent}};

\node[answer, below=0.5cm of p2, xshift=1.5cm] (ans) {\textbf{Final answer:} ``Dottie Hinson''\\[-2pt] {\scriptsize (fails without the specific fact)}};

\draw[arr] (q.south) -- ++(0,-0.2) -| (sq1.north);
\draw[arr] (sq1.south) -- (p1.north) node[midway, right, lbl] {retrieve};
\draw[arr] (p1.east) -- (v1.west);
\draw[arr] (p1.south) -- (sq2.north) node[midway, right, lbl] {next hop};
\draw[arr] (sq2.south) -- (p2.north) node[midway, right, lbl] {retrieve};
\draw[arr] (p2.east) -- (v2.west);
\draw[arr] (p2.south) -- ++(0,-0.2) -| (ans.north);

\node[right=0.2cm of v1, font=\scriptsize, text=green!40!black, text width=2.5cm] {Gold passage present, fact extractable};
\node[right=0.2cm of v2, font=\scriptsize, text=red!40!black, text width=2.5cm] {\textbf{Fact-grounding gap:} gold passage present but fact not stated};

\end{tikzpicture}
\caption{A 2-hop question from MuSiQue illustrating the \emph{fact-grounding gap}. At Hop 1, the retrieved passage contains the needed fact. At Hop 2, the gold passage is retrieved but does not state the character name. Standard retrieval metrics would mark both hops as successful; only Hop 1 actually contains the needed fact.}
\label{fig:example}
\end{figure*}

However, whether retrieved passages actually contain the specific facts needed at each reasoning step has not been systematically studied. Prior work has shown that irrelevant \citep{shi2024irrelevant} and distracting \citep{yoran2024making} context degrades QA performance, and recent frameworks decompose RAG errors into retriever and generator components \citep{ru2024ragchecker}. But these analyses operate at the document level --- they ask whether the right passage was retrieved, not whether a retrieved passage contains the relational fact the current reasoning step requires. This distinction has practical consequences: if a substantial fraction of failures stem from missing facts within retrieved documents rather than missing documents, then the widespread effort to improve retrievers addresses only part of the problem. This is particularly relevant as multi-hop QA systems are increasingly deployed in knowledge-intensive applications \citep{trivedi2023interleaving, lewis2020retrieval}, where misdiagnosing the source of failure leads to ineffective system improvements. Without this distinction, researchers risk investing in retrieval improvements that cannot address nearly half of the failures these systems encounter.

In this paper, we measure whether retrieved passages contain the specific facts each reasoning step requires, across three standard multi-hop QA benchmarks. We identify two distinct failure modes (illustrated in Figure~\ref{fig:example}). \emph{Retrieval failures} occur when the needed passage is not retrieved. \emph{Extraction failures} occur when the passage is retrieved but the needed fact cannot be extracted --- a phenomenon we term the \emph{fact-grounding gap}. Extraction failures account for as much as 47\% of all per-hop deficiencies. These failures are undetectable by standard retrieval metrics and persist across all retrieval interventions we test. To validate this finding, we train a lightweight fact-presence predictor, with an LLM judge validated against human annotation ($\kappa = 0.840$). We find targeted re-retrieval guided by this predictor matches blanket intervention at half the cost.

Our contributions are:
\begin{enumerate}
\item We identify the \emph{fact-grounding gap} --- a failure mode in which the correct passage is retrieved, but the required fact is not present. It appears across standard multi-hop QA benchmarks, accounting for 47\% of all per-hop deficiencies on MuSiQue.

\item We demonstrate that retrieval interventions, including augmentation and reranking, do not resolve extraction failures, establishing a ceiling for retrieval-only approaches.

\item We develop a per-hop fact-presence evaluation methodology, with an LLM judge validated against human annotation ($\kappa = 0.840$), and show that targeted re-retrieval on flagged hops matches every-hop intervention at half the cost.
\end{enumerate}

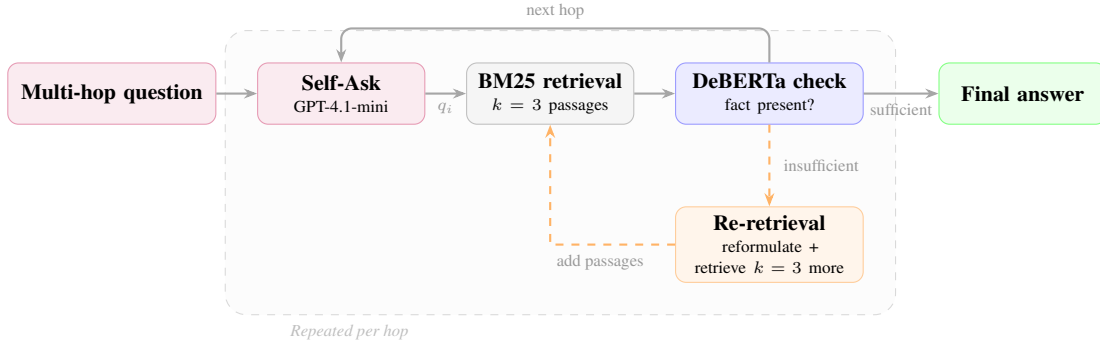
\begin{figure*}[t]
\centering
\begin{tikzpicture}[
    scale=0.9, transform shape,
    node distance=0.6cm and 0.5cm,
    box/.style={draw, rounded corners=4pt, minimum height=0.9cm, align=center, font=\small, text width=#1},
    box/.default=2.5cm,
    arr/.style={-{Stealth[length=5pt]}, thick, gray!70},
    darr/.style={-{Stealth[length=5pt]}, thick, dashed, orange!60},
    lbl/.style={font=\scriptsize, text=gray!80},
]
\node[box=2.8cm, fill=purple!8, draw=purple!40] (question) 
    {\textbf{Multi-hop question}};
\node[box=2.2cm, fill=purple!8, draw=purple!40, right=0.6cm of question] (selfask) 
    {\textbf{Self-Ask}\\[-1pt]{\scriptsize GPT-4.1-mini}};
\node[box=2.2cm, fill=gray!8, draw=gray!50, right=0.6cm of selfask] (retrieve) 
    {\textbf{BM25 retrieval}\\[-1pt]{\scriptsize $k=3$ passages}};
\node[box=2.5cm, fill=blue!8, draw=blue!40, right=0.6cm of retrieve] (deberta) 
    {\textbf{DeBERTa check}\\[-1pt]{\scriptsize fact present?}};
\node[box=2.2cm, fill=green!8, draw=green!50, right=1.1cm of deberta] (answer) 
    {\textbf{Final answer}};
\draw[arr] (question) -- (selfask);
\draw[arr] (selfask) -- (retrieve) node[midway, below, lbl] {$q_i$};
\draw[arr] (retrieve) -- (deberta);
\draw[arr] (deberta) -- (answer) node[midway, below, lbl] {sufficient};
\node[box=2.5cm, fill=orange!8, draw=orange!40, below=1.2cm of deberta] (reretrieval) 
    {\textbf{Re-retrieval}\\[-1pt]{\scriptsize reformulate +}\\[-1pt]{\scriptsize retrieve $k=3$ more}};
\draw[darr] (deberta.south) -- (reretrieval.north) node[midway, right, lbl, xshift=2pt] {insufficient};
\draw[darr] (reretrieval.west) -| (retrieve.south) node[pos=0.3, below, lbl] {add passages};
\draw[arr, rounded corners=4pt] (deberta.north) -- ++(0,0.5) -| (selfask.north) node[pos=0.25, above, lbl] {next hop};
\begin{scope}[on background layer]
\node[draw=gray!30, dashed, rounded corners=8pt, fill=gray!2,
      fit=(selfask)(retrieve)(deberta)(reretrieval),
      inner sep=12pt, 
      label={[font=\scriptsize\itshape, text=gray!50]below left:Repeated per hop}] {};
\end{scope}
\end{tikzpicture}
\caption{System architecture with BM25 retriever. At each hop, the Self-Ask pipeline decomposes the question, retrieves passages, and checks fact presence with the DeBERTa predictor. If flagged as insufficient, supplementary passages are retrieved before proceeding.}
\label{fig:architecture}
\end{figure*}

\section{Related Work}

\paragraph{Multi-hop question answering.}
Multi-hop QA requires combining evidence across multiple documents to answer complex questions. Early approaches used single-pass retrieval followed by reading comprehension \citep{yang2018hotpotqa}, but struggled with questions requiring evidence chains. More recent systems interleave retrieval with reasoning: IRCoT \citep{trivedi2023interleaving} chains retrieval with chain-of-thought prompting, Self-Ask \citep{press2023measuring} decomposes questions into explicit sub-questions with search calls, and ReAct \citep{yao2023react} combines reasoning and acting in a unified loop. These systems share a common pipeline --- decompose, retrieve, reason, repeat --- and a common assumption: that surfacing the right documents at each step is sufficient for correct reasoning. Our work does not propose a new pipeline but instead examines a failure mode common to all of them.

\paragraph{Improving multi-hop retrieval.}
Substantial work has focused on improving what gets retrieved at each hop. Approaches range from query decomposition and reformulation \citep{press2023measuring, trivedi2023interleaving} to cross-encoder reranking and adaptive retrieval strategies that decide when retrieval is necessary \citep{asai2024selfrag, jiang2023active, jeong2024adaptiverag}. These methods have meaningfully improved retrieval quality, but they share a common evaluation lens: success is measured by whether the right documents appear in the retrieved set. None examine whether a retrieved document contains the specific fact the current reasoning step requires --- the distinction our work is built on.

\paragraph{Evaluating retrieval in QA.}
Standard retrieval evaluation uses document-level metrics such as recall, precision, and mean reciprocal rank \citep{karpukhin2020dense}. Recent work has highlighted limitations of these metrics: \citet{yoran2024making} show that models are sensitive to irrelevant context, and \citet{shi2024irrelevant} demonstrate that adding irrelevant context degrades QA performance. \citet{zhu2025lost} identify a ``lost-in-retrieval'' problem where key entities are missed during sub-question decomposition, though their analysis operates at the entity level rather than the per-hop fact-presence level we examine. System-level failure decompositions distinguish retriever errors from reader errors, asking whether the retriever found the right passage or the reader extracted the wrong span. Our work operates at a finer granularity: we ask whether a retrieved passage --- even one that is topically relevant --- contains the specific relational fact needed for the current reasoning step.

\paragraph{LLM-based evaluation and annotation.}
Using LLMs as annotators and evaluators has become widespread for tasks where human annotation is expensive \citep{zheng2023judging, chiang2023can}. \citet{bavaresco2025llms} recommend validating LLM judges against task-specific human annotations before deployment. We follow this practice, validating our LLM fact-presence judge against human annotation ($\kappa = 0.840$) and documenting its conservative bias before using it to generate training labels for a lightweight classifier.

\section{Experimental Setup}
\paragraph{Datasets.} We evaluate on three standard multi-hop QA benchmarks: MuSiQue-Ans \citep{trivedi2022musique}, HotpotQA \citep{yang2018hotpotqa}, and 2WikiMultihopQA \citep{ho2020constructing}. We develop and validate our methodology on MuSiQue, which provides gold question decompositions with per-hop sub-questions, intermediate answers, and supporting paragraph indices, and apply the same procedure to HotpotQA and 2WikiMultihopQA.
\paragraph{QA system.} We use a Self-Ask style pipeline \citep{press2023measuring} with GPT-4.1-mini (\texttt{gpt-4.1-mini-2025-04-14}) as the reasoning model. At each hop, the model either generates a sub-question and retrieves $k=3$ passages via BM25 over Elasticsearch, or produces a final answer. Retrieved passages accumulate across hops and are deduplicated by content. To test whether the fact-grounding gap is retriever-dependent, we additionally run the full pipeline with a dense retriever, Contriever \citep{izacard2022contriever}, on MuSiQue and HotpotQA (\S\ref{sec:retriever_robustness}).

\paragraph{Evaluation.} We evaluate answer correctness using GPT-4.1-mini as an LLM judge \citep{zheng2023judging}. Unlike exact match, which penalizes correct answers with different surface forms (e.g., ``Christopher Nolan'' vs.\ ``Nolan''), the LLM judge handles synonyms, verbose answers, and partial overlaps. We validate this judge against human annotation on 200 stratified examples, achieving Cohen's $\kappa = 0.840$ (\citealt{landis1977measurement}).

\section{The Fact-Grounding Gap}

Standard retrieval evaluation in multi-hop QA measures whether the QA system retrieves the gold passage for each hop. We ask a finer question: does the retrieved passage actually contain the fact needed at the hop? In this section, we define and measure this distinction across 6,404 hops on MuSiQue, identifying two failure modes: retrieval failures, where the gold passage is missing, and extraction failures, where the gold passage is present but the needed fact is not.

\subsection{Measuring Per-Hop Fact Presence}
\label{sec:fact_presence_measurement}

Multi-hop questions decompose into a sequence of sub-questions, each requiring an intermediate answer that feeds into the next step, as illustrated in Figure~\ref{fig:example}. For each sub-question, we evaluate whether the accumulated retrieved passages contain the information needed to answer it. We collect all passages retrieved up to the current hop, remove duplicates, and pass them with the sub-question to an LLM judge. Sub-questions in shorthand format are normalized to natural language prior to evaluation (Appendix~\ref{app:normalization}).

We use GPT-4.1-mini as the fact-presence judge, following prior work on LLM-based evaluation \citep{zheng2023judging,chiang2023can}. The expected intermediate answer is withheld to prevent answer leakage. The judge outputs \textsc{Answerable} or \textsc{Not-Answerable} with a brief rationale, prompted with few-shot examples covering direct matches, implied answers, entity mentions without the required relational fact, and missing facts (Appendix~\ref{app:judge_prompt}).

A simpler alternative is string matching: checking whether the intermediate answer appears in the retrieved text. However, string matching systematically overstates fact presence \citep{maynez2020faithfulness}. On MuSiQue, this accounts for a 10.7-point gap between string matching (59.8\%) and LLM-judge answerability (49.1\%).

We validate the judge against human annotation on 200 stratified examples (100 per class), achieving 92.0\% agreement ($\kappa = 0.840$; \citealt{landis1977measurement}). Disagreements skew conservative: the judge more often marks hops as \textsc{Not-Answerable} when humans say \textsc{Answerable} (12 of 16 cases), so the reported gap may be slightly overstated. We additionally validate against gold paragraph annotations (Appendix~\ref{app:gold_consistency}).

\subsection{Results}
\label{sec:results}
Our per-hop measurements on MuSiQue yield two key findings.

\paragraph{Fact presence is low and degrades with hop depth.}
Across 6,404 hop-level evaluation steps on MuSiQue (2,417 questions), only 49.1\% are judged answerable from the system's retrieved passages. As shown in Table~\ref{tab:per_hop}, fact presence degrades sharply with reasoning depth: from 68.5\% at hop~1 to 18.5\% at hop~4. Later hops ask about entities that only become known during earlier reasoning steps, so their supporting evidence cannot be retrieved ahead of time. This degradation alone, however, does not reveal whether the needed passage was not retrieved, or was retrieved but lacked the required fact.

\begin{table}[h]
\centering
\small
\begin{tabular}{lcc}
\toprule
\textbf{Hop} & \textbf{Answerable (\%)} & \textbf{N} \\
\midrule
1 & 68.5 & 2{,}417 \\
2 & 43.7 & 2{,}417 \\
3 & 30.7 & 1{,}165 \\
4 & 18.5 & 405 \\
\midrule
All & 49.1 & 6{,}404 \\
\bottomrule
\end{tabular}
\caption{Per-hop fact-grounded answerability on MuSiQue dev.}
\label{tab:per_hop}
\end{table}

\paragraph{Failures decompose into two distinct modes.}
To distinguish whether failures arise from missing passages or from missing facts within retrieved passages, we cross-reference gold paragraph retrieval status with per-hop fact-presence labels from our LLM judge, yielding three categories over 6,401 hops.\footnote{Three hops are excluded due to missing gold paragraph annotations.} As shown in Table~\ref{tab:decomposition}, retrieval failures and extraction failures are nearly equally prevalent. Retrieval failures (30.7\%) occur when the gold supporting passage is not among the retrieved passages, a failure mode that stronger retrievers or re-retrieval can potentially address. Extraction failures (27.3\%) occur when the gold passage \emph{is} retrieved but does not contain the fact the reasoning step requires. This second failure mode is the more concerning: standard retrieval metrics such as document recall would mark these hops as successful, since the correct passage was retrieved. Yet the system still lacks the evidence it needs. As we show in \S\ref{sec:intervention}, no retrieval intervention we test recovers a fact missing from the retrieved passage; even re-retrieval can only help when a different passage in the corpus explicitly contains the fact. This indicates that the bottleneck is not retrieving better passages, but the absence of the required fact in the retrieved passage text itself. 

\begin{table}[h]
\centering
\small
\begin{tabular}{lcc}
\toprule
\textbf{Category} & \textbf{Count} & \textbf{\%} \\
\midrule
Retrieval failure & 1{,}962 & 30.7 \\
Extraction failure & 1{,}749 & 27.3 \\
Fact present & 2{,}690 & 42.0 \\
\bottomrule
\end{tabular}
\caption{Three-way decomposition of per-hop outcomes on MuSiQue dev (6,401 hops).}
\label{tab:decomposition}
\end{table}

Extraction failures account for 47\% of all per-hop deficiencies. This means that even a perfect retriever --- one that always retrieves the gold passage --- would leave nearly half of per-hop failures unresolved. Hop depth degradation in Table~\ref{tab:per_hop} appears to implicate both extraction and retrieval failure modes. However, extraction failures hold around 9\% at every hop depth, so the sharp decline in fact presence comes from retrieval failures alone.

Retrieval recall alone cannot distinguish between the two failure modes: among hops where the system answers incorrectly, recall and fact presence are nearly uncorrelated ($\rho = 0.042$). The best non-oracle predictor (retrieval recall threshold) reaches only 68.1\% accuracy, leaving a 15-point gap to oracle baselines (Table~\ref{tab:baselines}; full analysis in Appendix~\ref{sec:gap_metrics}).

\begin{table}[h]
\centering
\small
\begin{tabular}{lccc}
\toprule
\textbf{Method} & \textbf{Acc.\ (\%)} & \textbf{F1} & \textbf{Gold?} \\
\midrule
Majority & 50.9 & 0.000 & No \\
Hop number & 64.9 & 0.596 & No \\
Retrieval recall & 68.1 & 0.716 & No \\
\midrule
String match & 80.8 & 0.823 & Yes \\
Gold para.\ present & 83.4 & 0.834 & Yes \\
\bottomrule
\end{tabular}
\caption{Baseline fact-presence predictors on MuSiQue dev (6,404 hops). Methods below the divider use gold annotations.}
\label{tab:baselines}
\end{table}

\subsection{Robustness to Retriever Choice}
\label{sec:retriever_robustness}

The failure-mode decomposition in \S\ref{sec:results} uses BM25. To test whether extraction failures reflect a weakness of sparse retrieval rather than a property of the corpus, we re-run the full pipeline with Contriever \citep{izacard2022contriever}, a dense retriever, on all 2{,}417 MuSiQue dev questions and 1{,}000 HotpotQA bridge questions.

Extraction failures do not decrease under dense retrieval. Under a matched per-hop gold-retrieval criterion, extraction failures increase from 598 (BM25) to 725 (Contriever) on MuSiQue; they persist under the accumulated-passage criterion used in Table~\ref{tab:decomposition} as well. On hops where both retrievers retrieve the gold passage, 77\% of BM25's extraction failures are also extraction failures under Contriever, indicating these failures are properties of the corpus rather than of any particular retriever. Overall, MuSiQue accuracy is nearly unchanged (51.2\% vs.\ 51.9\%), and the same pattern holds on HotpotQA (3.1\% vs.\ 4.5\% extraction failures).

\section{Learning to Predict Fact-Presence Deficiency}
\label{sec:predictor}

Section 4 showed that retrieval scores cannot reliably predict fact presence. We train a lightweight classifier that predicts whether a hop's retrieved passages contain the needed fact, using only the sub-question and passages as input. The LLM judge from \S\ref{sec:fact_presence_measurement} can assess this accurately, but must evaluate each hop individually, making it slow and expensive to apply across thousands of hops (and use as part of the QA pipeline). Our classifier learns from the LLM judge's labels but is a much smaller model, making per-hop prediction practical.

\subsection{Training Setup}

We fine-tune DeBERTa-v3-large \citep{he2021debertav3} as a binary classifier predicting \textsc{Answerable} or \textsc{Not-Answerable} for each hop. The classifier takes the sub-question and accumulated retrieved passages as input, truncated to 1,500 characters. Training data consists of 46,610 MuSiQue hop-level examples labeled by the LLM judge described in \S\ref{sec:fact_presence_measurement} (53.9\% \textsc{Answerable}, 46.1\% \textsc{Not-Answerable}). 6,404 held-out examples from the MuSiQue dev set are used for evaluation. We use learning rate $2 \times 10^{-5}$, batch size 16, maximum sequence length 512, and 10\% linear warmup. Checkpoints are selected by dev F1 with early stopping (patience 3); training converges at epoch 3. At inference, we classify hops with predicted probability $\geq 0.5$ as \textsc{Answerable}. We develop the predictor on MuSiQue and evaluate its transfer to HotpotQA and 2WikiMultihopQA in intervention experiments (\S\ref{sec:intervention}).

\subsection{Classification Performance}

Table~\ref{tab:deberta_results} compares BERT-base, RoBERTa-large, and DeBERTa-v3-large on MuSiQue dev. All three substantially outperform the retrieval-recall baseline (68.1\%) from Section 4, with DeBERTa-v3-large achieving the highest F1 (0.785). This confirms that detecting whether the needed fact is present in a hop’s retrieved passages is a learnable task that generalizes across model architectures.

\begin{table}[h]
\centering
\small
\begin{tabular}{lcccc}
\toprule
\textbf{Model} & \textbf{Params} & \textbf{Prec.} & \textbf{Rec.} & \textbf{F1} \\
\midrule
BERT-base & 110M & 0.753 & 0.737 & 0.745 \\
RoBERTa-large & 355M & 0.840 & 0.683 & 0.754 \\
DeBERTa-v3-large & 435M & 0.805 & 0.765 & \textbf{0.785} \\
\bottomrule
\end{tabular}
\caption{Fact-presence predictor comparison on MuSiQue dev (6,404 hops).}
\label{tab:deberta_results}
\end{table}

The DeBERTa predictor flags 3,201 of 6,404 hops (50.0\%) as \textsc{Not-Answerable}, closely matching the ground-truth rate of 50.9\%. The close agreement between predicted and actual flag rates suggests the predictor can replace the LLM judge for identifying deficient hops, at a fraction of the computational cost. 

\paragraph{Label-quality and evidence-source ablation.}
We run ablations to test two potential bottlenecks in fact-presence prediction: the quality of the training labels, and the content of the retrieved passages. First, we replace LLM-judge labels with string-match labels, a straightforward alternative that marks a hop as answerable when the intermediate answer appears in the retrieved text. Second, we remove the retrieved passages from the input, forcing the predictor to rely on the sub-question alone.

Table~\ref{tab:ablation} shows the effect of each change. Switching from string-match labels to LLM-judge labels improves accuracy by 8.9 points, indicating that the presence of the answer string in the text does not reliably indicate whether the passage contains the needed fact. Adding the retrieved passages to the input yields a further 15.8-point improvement, confirming that the classifier needs to read the passage content to assess fact presence --- the sub-question alone is not enough. 

\begin{table}[h]
\centering
\small
\begin{tabular}{lcc}
\toprule
\textbf{Condition} & \textbf{Acc.\ (\%)} & \textbf{F1} \\
\midrule
String-match labels, no passages & 54.7 & 0.547 \\
LLM-judge labels, no passages & 63.6 & 0.615 \\
LLM-judge labels, with passages & 79.4 & 0.785 \\
\bottomrule
\end{tabular}
\caption{Ablation on label quality and input content.}
\label{tab:ablation}
\end{table}

These results justify our design choices: LLM-judge labels and the retrieved passages themselves are both necessary for accurate prediction. They also reinforce the finding from \S\ref{sec:results} that retrieval scores miss fact-presence information that is present in the passage text itself.

\paragraph{Flag count predicts QA failure.}
If the predictor is capturing a real distinction between fact-present and fact-deficient hops, then questions with more flagged hops should produce worse final answers. Table~\ref{tab:degradation} confirms this: accuracy drops steadily from 68.7\% with zero flagged hops to 8.5\% with four.  This consistent pattern shows that when more hops lack their needed fact, the QA system is less likely to produce the correct final answer.

\begin{table}[h]
\centering
\small
\begin{tabular}{lcc}
\toprule
\textbf{Flagged hops} & \textbf{N} & \textbf{Accuracy (\%)} \\
\midrule
0 & 556 & 68.7 \\
1 & 868 & 52.9 \\
2 & 690 & 44.9 \\
3 & 255 & 38.8 \\
4 & 47 & 8.5 \\
\bottomrule
\end{tabular}
\caption{QA accuracy by number of flagged hops on MuSiQue dev.}
\label{tab:degradation}
\end{table}

\section{Intervention Experiments}
\label{sec:intervention}

\begin{table*}[t]
\centering
\small
\begin{tabular}{llccc}
\toprule
\textbf{Dataset} & \textbf{Condition} & \textbf{Acc.\ (\%)} & \textbf{Flag\%} & \textbf{Calls/Q} \\
\midrule
\multirow{4}{*}{\textbf{MuSiQue} (BM25, Augment)} 
  & Baseline         & 51.9 & 0.0   & 0.00 \\
  & Always           & 54.6 & 100.0 & 2.65 \\
  & DeBERTa          & \textbf{55.6} & 50.0  & 1.32 \\
  & Oracle           & 54.9 & 48.9  & 1.29 \\
\midrule
\multirow{3}{*}{\textbf{MuSiQue} (BM25, Rerank)}
  & Baseline         & 51.8 & 0.0   & 0.00 \\
  & Always           & \textbf{60.9} & 100.0 & 2.65 \\
  & DeBERTa          & 60.3 & 50.0  & 1.33 \\
\midrule
\multirow{3}{*}{\textbf{MuSiQue} (Contriever, Augment)}
  & Baseline         & 51.2 & 0.0  & 0.00 \\
  & DeBERTa          & \textbf{54.7} & 45.2 & 1.20 \\
  & Oracle           & 54.8 & 50.9 & 1.35 \\
\midrule
\multirow{4}{*}{\textbf{HotpotQA} (BM25, Augment)}
  & Baseline         & 48.1 & 0.0   & 0.00 \\
  & Always           & \textbf{57.1} & 100.0 & 2.00 \\
  & DeBERTa          & 53.2 & 30.7  & 0.61 \\
  & Oracle           & 55.5 & 46.9  & 0.94 \\
\midrule
\multirow{3}{*}{\textbf{HotpotQA} (Contriever, Augment)}
  & Baseline         & 42.2 & 0.0  & 0.00 \\
  & DeBERTa          & \textbf{52.2} & 53.4 & 1.07 \\
  & Oracle           & 52.3 & 54.1 & 1.08 \\
\midrule
\multirow{3}{*}{\textbf{2WikiMQA} (BM25, Augment)}
  & Baseline         & 78.8 & 0.0   & 0.00 \\
  & Always           & 78.7 & 100.0 & 2.15 \\
  & DeBERTa          & \textbf{78.9} & 54.1  & 1.16 \\
\bottomrule
\end{tabular}
\caption{Intervention results across three datasets and two retrievers. Flag\%: percentage of hops receiving intervention. Calls/Q: average additional retrieval calls per question. 2WikiMQA and HotpotQA (Contriever) use dataset-specific retrained DeBERTa predictors; all other rows use the MuSiQue-trained predictor.}
\label{tab:main_results}
\end{table*}

In \S\ref{sec:predictor}, we showed that a classifier can reliably detect which hops lack their needed fact. We now test whether acting on these predictions improves QA accuracy by re-retrieving passages at flagged hops.

If the two failure modes identified in \S4 are genuinely distinct, we would expect re-retrieval to help when the needed passage was not retrieved. On the other hand, when the passage is retrieved but lacks the needed fact, re-retrieval should have negligible effect.
 We test these predictions using two retrieval interventions --- augmentation and reranking --- applied selectively at flagged hops, at every hop, or not at all.

\subsection{Experimental Setup}
\label{sec:setup}

\paragraph{Base QA pipeline.}
We use the same Self-Ask pipeline and retrieval setup described in \S3, applied to all three datasets.

\paragraph{Intervention policy.}
When the predictor flags a hop as absent of its needed fact, we attempt to find better evidence. We look at two ways of doing this:

\begin{itemize}
\item \textbf{Augmentation:} rephrase the sub-question and retrieve $k{=}3$ additional passages based on the rephrasing, appending them to the accumulated retrieved passages.
\item \textbf{Reranking:} retrieve a larger set of $k{=}20$ candidate passages, score them with a cross-encoder \citep{nogueira2019passage}, and keep the top 3.
\end{itemize}

\paragraph{Conditions and controls.}
We compare four conditions: \textbf{Baseline} (no intervention), \textbf{Always} (intervene at every hop), \textbf{DeBERTa} (intervene only at flagged hops), and \textbf{Oracle} (intervene only where the gold passage is actually missing). All conditions use the same retrieval corpus and reasoning model, differing only in which hops go through re-retrieval. This isolates the effect of the intervention. Statistical testing details are provided in Appendix~\ref{app:stats}.

\subsection{Main Results}
\label{sec:main_results}

Table~\ref{tab:main_results} presents intervention results across three datasets. On MuSiQue with augmentation, intervening only at hops the predictor flags as fact-deficient improves accuracy by +3.7\% over baseline, while intervening at every hop improves by +2.7\%. Despite using only half the re-retrieval calls, the predictor-guided approach performs as well as or better, indicating that it is identifying the hops where re-retrieval actually helps.

Reranking produces larger accuracy improvements than augmentation on MuSiQue, but in both cases, intervening only at flagged hops performs comparably (within 1\% accuracy) to intervening at every hop --- at half the re-retrieval cost. This is important because it shows the predictor's value is not tied to a specific intervention: the same pattern holds for both augmentation and reranking, despite the two working in fundamentally different ways. In both cases, the predictor correctly identifies which hops benefit from additional retrieval.

On HotpotQA, we apply the MuSiQue-trained predictor without retraining. It improves accuracy by +5.1\% over baseline while flagging only 30.7\% of hops for re-retrieval. Intervening at every hop yields a larger gain (+9.0\%), but this is expected for two-hop questions: with only two hops, unnecessary re-retrieval is unlikely to impact the answering process. On longer chain questions such as in MuSiQue, where questions have up to four hops, re-retrieving at every hop risks adding irrelevant passages that mislead the reasoning model at later steps. Selective intervention avoids this by re-retrieving only where the predictor identifies a genuine gap.

The targeted intervention also replicates with a dense retriever. We run the pipeline with Contriever on MuSiQue and HotpotQA, where re-retrieval improves accuracy by +3.5\% and +10.0\% respectively.

On 2WikiMultihopQA, neither augmentation nor reranking produces meaningful gains (baseline 78.8\%, DeBERTa 78.9\%). We discuss why in \S\ref{sec:entity_familiarity}.

All targeted intervention gains reported in this section --- across both BM25 and Contriever --- are statistically significant (McNemar's test, $p < 0.001$; Appendix~\ref{app:stats}).

\S\ref{sec:flag_degradation}--\ref{sec:failure_decomp} examine these results in more detail, breaking them down by the number of flagged hops, hop depth, and failure type.

\subsection{Flag-Count Degradation}
\label{sec:flag_degradation}

The results in \S\ref{sec:main_results} show that targeted intervention improves accuracy overall, but do not reveal whether the predictor is useful at the individual question level. If the predictor is accurately identifying fact-deficient hops, then questions with more flagged hops should be harder, and intervention should help those questions more.

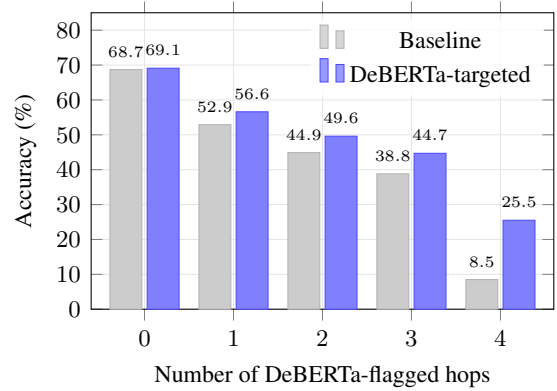
\begin{figure}[t]
\centering
\begin{tikzpicture}
\begin{axis}[
    ybar,
    bar width=12pt,
    width=\columnwidth,
    height=5.5cm,
    ylabel={Accuracy (\%)},
    xlabel={Number of DeBERTa-flagged hops},
    ymin=0, ymax=85,
    xtick={0,1,2,3,4},
    ytick={0,10,20,30,40,50,60,70,80},
    legend style={
        at={(0.98,0.98)},
        anchor=north east,
        font=\small,
        draw=none,
        fill opacity=0.8,
        text opacity=1,
    },
    nodes near coords,
    every node near coord/.append style={font=\tiny, above, yshift=1pt},
    enlarge x limits=0.15,
    grid=major,
    grid style={gray!20},
    tick label style={font=\small},
    label style={font=\small},
]
\addplot[fill=gray!40, draw=gray!60] coordinates {
    (0, 68.7)
    (1, 52.9)
    (2, 44.9)
    (3, 38.8)
    (4, 8.5)
};
\addplot[fill=blue!50, draw=blue!70] coordinates {
    (0, 69.1)
    (1, 56.6)
    (2, 49.6)
    (3, 44.7)
    (4, 25.5)
};
\legend{Baseline, DeBERTa-targeted}
\end{axis}
\end{tikzpicture}
\caption{QA accuracy by number of DeBERTa-flagged hops on MuSiQue (augmentation).}
\label{fig:degradation}
\end{figure}

Figure~\ref{fig:degradation} confirms both predictions. Questions with zero flagged hops achieve 68.7\% baseline accuracy, while questions with four flagged hops achieve only 8.5\%. The more hops the predictor flags, the worse the question performs without intervention. More importantly, the benefit of intervention grows with the number of flags: questions with zero flags gain almost nothing from re-retrieval (+0.4\%), while questions with four flags gain +17.0\%. The predictor is not just detecting harder questions --- it is identifying the questions where re-retrieval can actually help.

The same pattern appears on HotpotQA without any retraining: accuracy drops steadily as flag count increases, and the improvement from re-retrieval intervention grows with the number of flags (from +0.1\% at zero flags to +11.8\% at two). This suggests the predictor has learned a general signal about when retrieved passages lack the needed fact, not a dataset-specific pattern.

\subsection{Selectivity: When Targeting Matters}
\label{sec:selectivity}

The results in \S\ref{sec:main_results} show that targeted and every-hop intervention achieve similar overall accuracy. But if we break results down by the number of hops in each question, a clear difference emerges.

At 3-hop questions, intervening at every hop actually \emph{hurts} accuracy, dropping it by 2.1\% below baseline. Targeted intervention, by contrast, \emph{improves} accuracy by 2.1\% --- a 4.2-point difference between the two methods on the same set of questions. With three hops, re-retrieving everywhere means at least some hops receive unnecessary passages that mislead later reasoning steps. The predictor avoids this by leaving well-retrieved hops alone. 

At 2-hop questions, every-hop intervention performs better, and at 4-hop, both strategies produce similar large gains, likely because these questions are difficult enough that the remaining errors cannot be addressed by retrieval alone.

This pattern explains the overall results: every-hop intervention gains an advantage on 2-hop questions, but loses it on 3-hop questions where it actually hurts. Despite similar results at 4-hop, targeted intervention is the safer strategy: it achieves comparable accuracy across all hop depths, while every-hop intervention risks hurting accuracy in the cases where only some hops are fact-deficient.

\subsection{Failure Mode Decomposition}
\label{sec:failure_decomp}

The previous subsections show that re-retrieval improves accuracy overall but does not help equally across all different types of questions. A natural follow up to this is: which types of failures does re-retrieval actually fix? In \S4, we identified two failure modes --- retrieval failures (the needed passage was not retrieved) and extraction failures (the passage was retrieved but does not contain the needed fact). If these two failure modes are genuinely distinct, re-retrieval should improve accuracy on retrieval failures but not on extraction failures.

This is exactly what we find. Accuracy on retrieval failures improves substantially with both augmentation (+8.6\%) and reranking (+25.3\%), as re-retrieval can potentially retrieve the previously missing gold passage. Accuracy on extraction failures shows negligible change ($-0.5$\%). The extraction failure results especially make sense, as the gold passage had already been retrieved prior to re-retrieval.

This result has a direct practical implication: it establishes a ceiling on what retrieval improvements can achieve in multi-hop QA. No matter how effective the retriever becomes, extraction failures --- which account for 27.3\% of all hops --- will remain unaddressed.

This ceiling reflects genuine corpus gaps. For each extraction failure, we searched all corpus passages containing the answer string to check whether any states the needed relational fact. The fact is absent in the vast majority of cases (97.7\% MuSiQue, 100\% HotpotQA, 97.6\% 2WikiMultihopQA). The answer string often appears elsewhere in the corpus, but no passage states the relational fact the reasoning step requires.

\subsection{Entity Familiarity and Limits of Transfer}
\label{sec:entity_familiarity}

The gains on MuSiQue and HotpotQA raise a natural question: does the fact-grounding gap exist on other multi-hop QA datasets? We take a look at 2WikiMultihopQA:  neither augmentation nor reranking produces meaningful gains (baseline 78.8\%, DeBERTa 78.9\%). The baseline accuracy is already high, likely because 2WikiMultihopQA questions involve well-known entities with extensive Wikipedia coverage, so the initial retrieval is sufficient for most hops. The predictor flags 54.1\% of hops, yet re-retrieval at those hops produces no accuracy gain --- suggesting that most flags on this dataset do not correspond to genuine fact deficiencies. Retraining the predictor on 2WikiMultihopQA data does not help either (78.9\% vs 78.8\% baseline), confirming that the limitation is not a dataset transfer issue but rather that the aggregate gap on the dataset is small. The gap is present but concentrated: comparison questions show only 2.7\% extraction failures, while multi-step questions have 16.6\% extraction failures. This result suggests that the severity of the fact-grounding gap varies --- but is present --- across datasets and question types.

\section{Conclusion}

We introduced the \emph{fact-grounding gap} --- the phenomenon where multi-hop QA systems retrieve topically relevant documents that nonetheless lack the specific facts needed for each reasoning step. Through per-hop analysis across MuSiQue, HotpotQA, and 2WikiMultihopQA, we identified two distinct failure modes: retrieval failures and extraction failures. On MuSiQue, extraction failures account for 27.3\% of all hops and are invisible to standard retrieval metrics. No retrieval intervention we test --- augmentation or reranking --- recovers a fact missing from the retrieved passage, establishing a concrete ceiling for retrieval-only approaches.

These findings generalize: on HotpotQA, the same patterns --- declining accuracy with more flagged hops, and larger intervention gains on flagged questions --- appear without retraining the predictor. The fact-grounding gap is not specific to MuSiQue or to sparse retrieval, but a recurring property of multi-hop QA systems.

\section*{Limitations}

Our study has several limitations. First, our analysis uses a single reasoning model (GPT-4.1-mini). While the cross-dataset and cross-retriever transfer results suggest generalizability, the specific failure rates and intervention effects may differ with other reasoning models. Second, our LLM judge, despite strong human agreement ($\kappa = 0.840$), may exhibit systematic biases on question types not well-represented in the few-shot examples. Third, our interventions show substantial gains only on retrieval failures; we do not propose solutions for extraction failures, which likely require approaches beyond retrieval, such as improved passage representation or fact-aware reading comprehension. Fourth, we evaluate on English-language datasets with Wikipedia-based corpora; the prevalence of the fact-grounding gap in other languages, domains, or corpus types remains unexplored. Fifth, we evaluate the fact-presence predictor only in multi-hop QA; because this system requires just a query and its retrieved passages, it could apply to other retrieval-based pipelines such as retrieval-augmented generation, but we do not test that setting.

\clearpage
\bibliography{references}

\clearpage

\appendix

\section{LLM Judge Prompt}
\label{app:judge_prompt}

The fact-presence judge uses the following prompt template with few-shot examples:

\begin{quote}
\small
\texttt{You are evaluating whether retrieved passages contain enough information to answer a specific sub-question in a multi-hop reasoning chain.}\\[0.5em]
\texttt{Given a sub-question and retrieved passages, determine:}\\
\texttt{- YES: The passages contain enough information to answer the sub-question (even if the answer is implied, paraphrased, or requires minor inference).}\\
\texttt{- NO: The passages do NOT contain enough information to answer the sub-question. The needed fact is missing.}\\[0.5em]
\texttt{IMPORTANT: Judge whether the SPECIFIC FACT needed to answer this sub-question is present. A passage can be topically related but still lack the specific fact needed.}\\[0.5em]
\texttt{Sub-question: \{sub\_question\}}\\
\texttt{Passages: \{passages\}}\\
\texttt{Reasoning:}
\end{quote}

\noindent Example:

\begin{quote}
\small
\textbf{Sub-question:} Trey Parker: place of birth \\
\textbf{Passages:} [1] \emph{South Park}: South Park is an American animated sitcom created by Trey Parker and Matt Stone. The show premiered on August 13, 1997, on Comedy Central. \\
\textbf{Reasoning:} The passage mentions Trey Parker as a creator of South Park, but says nothing about where he was born. \\
\textbf{Verdict:} NO
\end{quote}

\noindent The complete set of six examples is included in our released code.

\section{Sub-Question Normalization Examples}
\label{app:normalization}

MuSiQue stores some sub-questions in a shorthand format derived from knowledge graph triples. We normalize these to natural language using GPT-4.1-mini (Table~\ref{tab:normalization_examples}).

\begin{table}[h]
\centering
\small
\begin{tabular}{p{3cm}p{3.5cm}}
\toprule
\textbf{Shorthand} & \textbf{Normalized} \\
\midrule
UHF $\gg$ distributed by & Which company distributed UHF? \\
Green $\gg$ performer & Who is the performer of Green? \\
Learjet 60 $\gg$ manufacturer & Who manufactures the Learjet 60? \\
Ciudad Deportiva $\gg$ owner & Who owns Ciudad Deportiva? \\
\bottomrule
\end{tabular}
\caption{Sub-question normalization examples.}
\label{tab:normalization_examples}
\end{table}

Of 6,404 sub-questions on MuSiQue dev, 4,100 are already in natural language; the remaining 2,304 require conversion (1,315 unique shorthand patterns). All conversions are saved as an auditable mapping file.

\section{Gold Paragraph Consistency}
\label{app:gold_consistency}

We compare LLM labels against MuSiQue's gold paragraph annotations. When the gold supporting paragraph is retrieved at a hop, the judge labels this hop as answerable 81.7\% of the time (2,677 of 3,275 hops). The remaining 18.3\% represent genuinely ambiguous cases where the gold paragraph does not self-sufficiently answer the sub-question. When the gold paragraph is not retrieved, the judge labels only 15.0\% as answerable (468 of 3,129 hops) --- these are cases where the needed fact appears in a non-gold passage.

\section{Retrieval Recall Correlation Analysis}
\label{sec:gap_metrics}

If retrieval quality alone explained extraction failures, hops with higher retrieval recall should show higher fact presence. When we consider all 6,404 hops on the MuSiQue dev set, retrieval recall and fact presence are moderately correlated (Spearman $\rho = 0.449$): hops where more gold passages are retrieved tend to be more answerable. However, this relationship largely reflects easy cases where retrieval and fact presence succeed or fail together. Among the 5,283 hops where the system answers incorrectly --- the cases where diagnosis matters most --- the correlation nearly vanishes ($\rho = 0.042$). Retrieval recall provides almost no signal about whether a failure is a retrieval problem or an extraction one.

Table~\ref{tab:baselines} (main text) compares several baselines for predicting per-hop fact presence. Without gold annotations, the best predictor (retrieval recall) reaches only 68.1\% accuracy. Two methods using gold annotations perform substantially better: string matching the intermediate answer against retrieved text (80.8\%) and verifying gold paragraph presence (83.4\%). These require knowing the gold answer in advance and cannot be applied at inference time. The 15-point gap between retrieval recall (68.1\%) and gold paragraph presence (83.4\%) indicates that fact presence is predictable from passage content, but standard retrieval metrics cannot capture it \citep{ru2024ragchecker}.

\section{Predictor Complementarity Analysis}
\label{app:complementarity}

We compare DeBERTa against a BM25 score threshold (flagging hops where the maximum retrieval score falls below the corpus median) and a random predictor (flagging each hop with probability 0.5). Table~\ref{tab:predictor_comparison} reports the Spearman correlation between the number of flagged hops per question and QA failure.

\begin{table}[h]
\centering
\small
\begin{tabular}{lcc}
\toprule
\textbf{Predictor} & \textbf{Spearman $\rho$} & \textbf{$p$-value} \\
\midrule
Random & $-0.066$ & $1.11 \times 10^{-3}$ \\
DeBERTa & $-0.216$ & $5.76 \times 10^{-27}$ \\
BM25-threshold & $-0.254$ & $6.54 \times 10^{-37}$ \\
DeBERTa $\cap$ BM25 & $\mathbf{-0.282}$ & $1.72 \times 10^{-45}$ \\
\bottomrule
\end{tabular}
\caption{Spearman correlation between flag count and QA failure. The intersection of DeBERTa and BM25 yields the strongest signal, as the two capture complementary failure dimensions.}
\label{tab:predictor_comparison}
\end{table}

BM25-threshold achieves a stronger individual correlation ($\rho = -0.254$) than DeBERTa ($\rho = -0.216$), as retrieval score directly reflects query--document match quality. However, the two signals capture complementary failure dimensions. At the hop level, DeBERTa and BM25-threshold flags overlap with a Jaccard coefficient of only 42.9\%, indicating that they identify substantially different sets of deficient hops.

\begin{table}[h]
\centering
\small
\begin{tabular}{lcc}
\toprule
\textbf{Category} & \textbf{Hops} & \textbf{\%} \\
\midrule
Both flag & 1{,}971 & 30.8 \\
DeBERTa-only & 1{,}230 & 19.2 \\
BM25-only & 1{,}391 & 21.7 \\
Neither & 1{,}809 & 28.3 \\
\bottomrule
\end{tabular}
\caption{Hop-level overlap between DeBERTa and BM25-threshold flags (6,401 hops). Jaccard overlap: 42.9\%.}
\label{tab:complementarity_hops}
\end{table}

To understand what each signal captures, we measure baseline QA accuracy for questions grouped by which predictor(s) flag at least one hop.

\begin{table}[h]
\centering
\small
\begin{tabular}{lcc}
\toprule
\textbf{Signal source} & \textbf{$n$} & \textbf{Baseline acc.\ (\%)} \\
\midrule
Neither flags & 214 & 76.6 \\
DeBERTa-only flags & 208 & 71.6 \\
BM25-only flags & 342 & 63.7 \\
Both flag & 1{,}434 & 41.0 \\
\bottomrule
\end{tabular}
\caption{Baseline QA accuracy by which predictor(s) flag the question. DeBERTa-only questions retain relatively high accuracy, suggestive of extraction-type deficiencies where retrieval appeared adequate. BM25-only questions have lower accuracy, suggestive of retrieval-risk cases.}
\label{tab:complementarity_acc}
\end{table}

Questions flagged only by DeBERTa exhibit relatively high baseline accuracy (71.6\%), suggestive of extraction-type deficiencies: the retrieval system found relevant documents (hence no BM25 flag), but the needed fact is absent or not extractable from the retrieved passages. Questions flagged only by BM25 have lower accuracy (63.7\%), suggestive of retrieval-risk cases: the retrieval scores are low because the system failed to find relevant documents. Their intersection produces the strongest predictor ($\rho = -0.282$, $p < 10^{-45}$), as the two signals capture complementary failure dimensions.

\section{Supervised Fine-Tuning Experiment}
\label{app:sft}

To test whether the fact-grounding signal can improve the reasoning model itself, we fine-tune GPT-4.1-mini on 654 successful Self-Ask reasoning traces --- questions where the baseline model produced correct answers. The training data consists of the exact multi-turn prompt-response pairs from the Self-Ask pipeline, with no intervention examples.

The fine-tuned model improves by +3.5 points over the base model (48.1\% $\to$ 51.6\%, LLM-judged on $n = 2{,}417$), comparable to the DeBERTa intervention gain (+3.7\%). The training data contains no intervention traces; the model learns better reasoning patterns purely from its own successful trajectories. SFT and DeBERTa-targeted intervention represent complementary approaches: one improves the evidence at inference time, the other improves reasoning at training time.

\section{Retrieval Corpus Details}
\label{app:corpora}

For each dataset, we index passage contexts in Elasticsearch using default BM25 scoring (Table~\ref{tab:corpora}). For the dense-retrieval experiments (\S\ref{sec:retriever_robustness}), we encode the MuSiQue and HotpotQA passage sets with Contriever (\texttt{facebook/contriever}) and index them with FAISS.

\begin{table}[h]
\centering
\small
\setlength{\tabcolsep}{4pt}
\resizebox{\columnwidth}{!}{%
\begin{tabular}{lrl}
\toprule
\textbf{Dataset} & \textbf{Passages} & \textbf{Source} \\
\midrule
MuSiQue & 139{,}416 & Train + dev passage contexts \\
HotpotQA & 507{,}494 & Full Wikipedia abstracts \\
2WikiMultihopQA & 136{,}009 & Dev + 20K train sample contexts \\
\bottomrule
\end{tabular}}
\caption{Retrieval corpus statistics per dataset.}
\label{tab:corpora}
\end{table}

\begin{table*}[!tb]
\centering
\small
\begin{tabular}{llllccc}
\toprule
\textbf{Dataset} & \textbf{Retriever} & \textbf{Intervention} & \textbf{Predictor} & \textbf{Gain} & \textbf{$p$} & \textbf{95\% CI} \\
\midrule
MuSiQue  & BM25       & Reranking    & MuSiQue-trained    & +8.5  & $5{\times}10^{-22}$   & $[+6.8, +10.2]$ \\
MuSiQue  & BM25       & Augmentation & MuSiQue-trained    & +3.7  & $3{\times}10^{-6}$    & $[+2.2, +5.2]$ \\
MuSiQue  & Contriever & Augmentation & MuSiQue-trained    & +3.5  & $8{\times}10^{-7}$    & $[+2.1, +4.8]$ \\
HotpotQA & BM25       & Augmentation & MuSiQue-trained    & +5.1  & $< 10^{-33}$          & $[+4.3, +5.9]$ \\
HotpotQA & Contriever & Augmentation & MuSiQue-trained    & +6.8  & $3.5{\times}10^{-10}$ & $[+4.8, +8.9]$ \\
HotpotQA & Contriever & Augmentation & HotpotQA-retrained & +10.0 & $3{\times}10^{-15}$   & $[+7.6, +12.4]$ \\
HotpotQA & BM25       & Augmentation & HotpotQA-retrained & +8.7  & $9{\times}10^{-14}$   & $[+6.5, +10.9]$ \\
\bottomrule
\end{tabular}
\caption{McNemar's test for targeted (DeBERTa) intervention vs.\ baseline. Gains and 95\% confidence intervals in percentage points. All gains significant at $p < 0.001$.}
\label{tab:significance}
\end{table*}

\section{Statistical Testing Details}
\label{app:stats}

For paired accuracy comparisons (e.g., DeBERTa vs.\ Baseline), we use McNemar's test on the $2 \times 2$ contingency table of per-question correct/incorrect outcomes. We report bootstrap 95\% confidence intervals for key effect sizes, computed with 10,000 resamples. Spearman rank correlations are used for monotonic associations (e.g., flag count vs.\ QA accuracy) because the outcome is ordinal and non-Gaussian. All reported $p$-values are two-sided unless otherwise noted. Per-comparison results are given in Table~\ref{tab:significance}.

\section{DeBERTa Training Details}
\label{app:training_details}

\paragraph{MuSiQue predictor.}
Model: DeBERTa-v3-large (435M parameters, 24 layers, 1,024 hidden dimensions). Training data: 46,610 hop-level examples from MuSiQue train. Dev set: 6,404 examples from MuSiQue dev. Learning rate: $2 \times 10^{-5}$. Batch size: 16. Max sequence length: 512. Warmup: 10\% linear. Early stopping: patience 3, selected by dev F1. Converged at epoch 3.

\paragraph{2WikiMultihopQA predictor.}
Same architecture retrained on 2WikiMultihopQA labels. Training data: hop-level examples from 2WikiMQA train trajectories, labeled by the same LLM-judge procedure. The retrained predictor achieves 97.4\% accuracy (F1 = 0.983) on 2WikiMQA dev, with bimodal probability distribution (flagged hops $\approx 0.000$, non-flagged $\approx 0.999$). Zero-shot transfer from MuSiQue failed on this dataset (0\% flag rate), necessitating domain-specific retraining.

\paragraph{HotpotQA predictor.}
Same architecture retrained on HotpotQA labels generated by the same LLM-judge procedure, with hyperparameters identical to the MuSiQue predictor.

\paragraph{BERT-base and RoBERTa-large.}
Trained with identical hyperparameters and data as the MuSiQue DeBERTa predictor. BERT-base: 110M parameters, 12 layers. RoBERTa-large: 355M parameters, 24 layers. Results reported in Table~\ref{tab:deberta_results} in main text.

\section{Compute Infrastructure}
\label{app:compute}

DeBERTa training was performed on NVIDIA A100 GPUs (PCIe and SXM variants) via RunPod, at a total cost of approximately \$80. The QA pipeline and all experiments were run on a 2-vCPU DigitalOcean VM (Ubuntu 24) with no GPU. LLM API calls used GPT-4.1-mini via the OpenAI API at a total cost of approximately \$100.

\section{Dataset and Model Licenses}
\label{app:dataset_model_license}

\paragraph{Datasets.}
MuSiQue-Ans is released under the CC BY 4.0 license. HotpotQA is under CC BY-SA 4.0. 2WikiMultihopQA is under Apache 2.0. All three datasets used per stated intended use.

\paragraph{Models.}
DeBERTa-v3-large is under the MIT License. BERT-base and RoBERTa-large are under Apache 2.0 and MIT, respectively. GPT-4.1-mini is accessed via the OpenAI API under OpenAI's Terms of Use. Our applications follow the intended use of these terms and applicable model cards. 

Contriever is under the CC BY-NC 4.0 license and FAISS under the MIT License. Our use of Contriever is non-commercial academic research, consistent with its license.

Elasticsearch is under the Server Side Public License (SSPL) v1. Our applications follow the intended use.
\end{document}